\documentclass{article}
\usepackage{arxiv}
\usepackage[utf8]{inputenc}
\usepackage{graphicx}
\graphicspath{ {./images/} }
\usepackage{booktabs}
\usepackage{caption} 
\usepackage{subcaption}
\usepackage{algorithm}
\usepackage{algorithmic}
\usepackage{amsmath}
\usepackage[dvipsnames]{xcolor}

\title{A Hardware-Aware System for Accelerating Deep Neural Network Optimization}

\author{
    Anthony Sarah \\
    \small{Intel Labs, Intel Corporation} \\
    \small{anthony.sarah@intel.com} \\\And
    Daniel Cummings \\
    \small{Intel Labs, Intel Corporation} \\
    \small{daniel.cummings@intel.com} \\\And
    Sharath Nittur Sridhar \\
    \small{Intel Labs, Intel Corporation} \\
    \small{sharath.nittur.sridhar@intel.com} \\\And
    Sairam Sundaresan \\
    \small{Intel Labs, Intel Corporation} \\
    \small{sairam.sundaresan@intel.com} \\\And
    Maciej Szankin \\
    \small{Intel Labs, Intel Corporation} \\
    \small{maciej.szankin@intel.com} \\\And
    Tristan Webb \\
    \small{Intel Labs, Intel Corporation} \\
    \small{tristan.webb@intel.com} \\\And
    J. Pablo Mu\~{n}oz \\
    \small{Intel Labs, Intel Corporation} \\
    \small{Pablo.munoz@intel.com}
}

\date{}

\begin{document}

\maketitle

\begin{abstract}
Recent advances in Neural Architecture Search (NAS) which extract specialized hardware-aware configurations (a.k.a. "sub-networks") from a hardware-agnostic "super-network" have become increasingly popular. While considerable effort has been employed towards improving the first stage, namely, the training of the super-network, the search for derivative high-performing sub-networks is still largely under-explored. For example, some recent network morphism techniques allow a super-network to be trained once and then have hardware-specific networks extracted from it as needed. These methods decouple the super-network training from the sub-network search and thus decrease the computational burden of specializing to different hardware platforms. We propose a comprehensive system that automatically and efficiently finds sub-networks from a pre-trained super-network that are optimized to different performance metrics and hardware configurations. By combining novel search tactics and algorithms with intelligent use of predictors, we significantly decrease the time needed to find optimal sub-networks from a given super-network. Further, our approach does not require the super-network to be refined for the target task a priori, thus allowing it to interface with any super-network. We demonstrate through extensive experiments that our system works seamlessly with existing state-of-the-art super-network training methods in multiple domains. Moreover, we show how novel search tactics paired with evolutionary algorithms can accelerate the search process for ResNet50, MobileNetV3 and Transformer while maintaining objective space Pareto front diversity and demonstrate an 8x faster search result than the state-of-the-art Bayesian optimization WeakNAS approach.
\end{abstract}

\section{Introduction}

Artificial intelligence (AI) researchers are continually pushing the state-of-the-art by creating new deep neural networks (DNNs) for different application domains (e.g., computer vision, natural language processing). In many cases, the DNNs are created and evaluated on the hardware platform available to the researcher at the time (e.g., GPU). Furthermore, the researcher may have only been interested in a narrow set of performance metrics such as accuracy when evaluating the network. Therefore, the network is inherently optimized for a specific hardware platform and specific metrics.

However, users wanting to solve the same problem for which the network was designed may have different hardware platforms available and may be interested in different and/or multiple performance metrics (e.g., accuracy \textit{and} latency). The performance of the network provided by the researcher is then suboptimal for these users.

Unfortunately, optimizing the network for the user's hardware and performance metrics (a.k.a. objectives) is a time-consuming effort requiring highly specialized knowledge. Network optimization is typically done manually with a great deal of in-depth understanding of the hardware platform since certain hardware characteristics (e.g., clock speed, number of logical processor cores, amount of cache memory, amount of RAM) will affect the optimization process. The optimization process is also affected by the characteristics of the input data to the DNN (e.g., batch size, image size). Finally, any change to the performance objectives (e.g., going from latency to power consumption), input data characteristics (e.g., increasing the batch size), hardware characteristics (e.g., changing the number of dedicated logical cores) or hardware platform (e.g., going from GPU to CPU) would require starting this expensive optimization process again. With this in mind, we adopt a "super-network" approach for addressing the hardware-aware model optimization task. This neural architecture search (NAS) approach generates a highly-diverse set of architectural options (a.k.a. sub-networks) for a reference architecture and allows us to exploit an extremely large search space to find optimal models for a particular hardware and objective setting. Additionally, this approach offers insight into what constitutes an optimal model for a given hardware setting. 

\begin{figure*}
    \center{\includegraphics[width=1.0\linewidth] {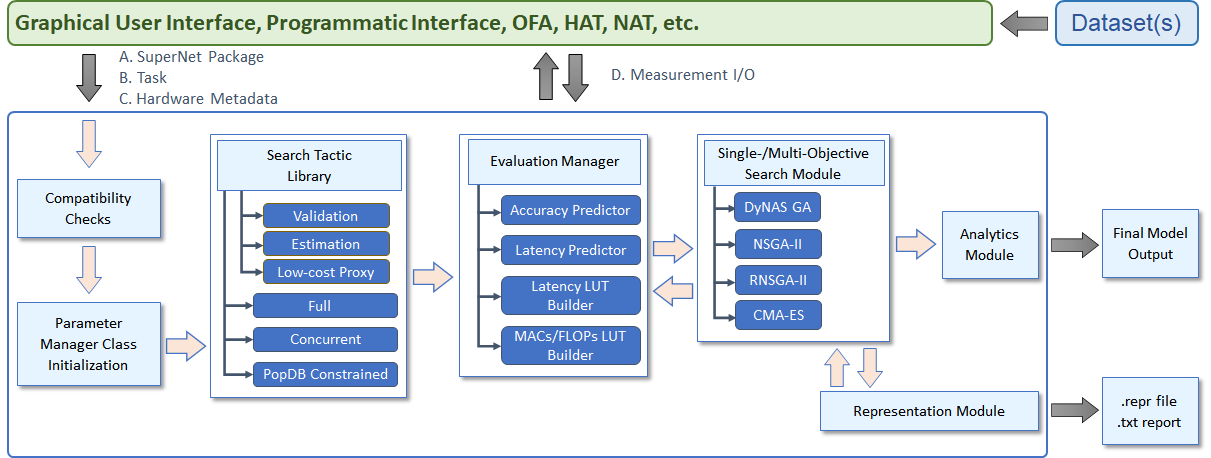}}
    \caption{\label{fig:system_overview} Our system is able to interface graphically or programmatically with users or through NAS systems such as OFA to provide them with hardware-optimized DNN architectures. The user simply specifies the data set, super-network, task and hardware metadata through these interfaces to begin the search. A number of novel search tactics, evaluation strategies and search algorithms are provided which not only find high-performing DNN architectures (e.g., low latency and high accuracy) but does so with high algorithmic efficiency (i.e., low number of search iterations). }
\end{figure*}

We propose an overall system design (Figure \ref{fig:system_overview}) which allows users to automatically and efficiently find networks that are optimized for their hardware platform. This design also works jointly with any existing or future super-network framework. In this work, demonstrate its application in the image classification and machine translation domains using the Once-For-All (OFA) \cite{cai2019once} and Hardware-Aware Transformers (HAT) \cite{wang2020hat} super-network frameworks respectively. Our solution provides DNN architectures applicable to specified application domains and optimized for specified hardware platforms in significantly less time than could be done manually. 

In this work, our primary contributions (1) demonstrate a modular and flexible system for accelerating sub-network discovery for any super-network framework, (2) examine why models need to be re-optimized based on the hardware platform, (3) show the efficiency of evolutionary algorithms for generating a diverse set of models, (4) demonstrate an accelerated approach to finding optimal sub-networks termed \textit{ConcurrentNAS}, and (5) propose a novel unsupervised methodology to automatically reduce the search space termed \textit{PopDB}.

\section{Related Work}

Many state-of-the-art neural architecture search (NAS) methods focus on decreasing the time required to find optimal models. Some of these approaches use the concept of a super-network, that is, a structure from which smaller sub-networks can be extracted. A recent approach, OFA, significantly reduces the time required for the search stage by decoupling training and search. Efficient mechanisms, such as accuracy, latency, and FLOPs predictors are used to speed up the search, avoiding the need for evaluating large portions of the model space which, depending on the sample size, is often unfeasible. The main focus of OFA was on computer vision architectures (MobileNetV3, ResNet50). 

There have been attempts from the same research group to extend the OFA approach to Natural Language Processing (NLP). Hardware-aware Transformers (HAT) \cite{wang2020hat} achieve this goal by extending network elasticity to this domain. In HAT, the authors introduce \textit{arbitrary encoder-decoder attention}, to break the information bottleneck between the encoder and decoder layers in Transformers \cite{vaswani2017attention}. Additionally, they propose \textit{heterogenous transformer layers} to allow for different layers to have different parameters. Using these techniques for constructing the design space, HAT discovers efficient models for different hardware platforms. 

The main limiting factor to improving sub-network search is the size of the model space which is defined by the \textit{elastic parameters} (i.e., the super-network parameters which vary in sub-networks) and the values they can take. There are two themes in research to address this problem: find ways to improve the search time and find ways to reduce the search space complexity. SemiNAS uses a encoder-predictor-decoder framework with LSTMs to predict the accuracy of architectures but is limited to a single-objective context \cite{Luo2020}. LaNAS recursively partitions the search space based on the sub-network performance \cite{wang2021sample}. Recently, WeakNAS \cite{wu2021stronger} demonstrated that SemiNAS and LaNAS give sub-optimal performance and showed how Bayesian optimization paired with weak predictors could be used to accelerate the sub-network search. In the results section we benchmark against WeakNAS to demonstrate the efficiency of our system. Many works such as OFA and WeakNAS take the approach of first training predictors and followed by search where an optional fine-tuning phase of the discovered sub-networks is used to push those models to achieve state-of-the-art performance. Neural Architecture Transfer (NAT) takes a creative approach to training super-networks by using genetic algorithms to inform configurations of the super-network on which to focus training \cite{Lu_2021_NAT} and show benefits over the related NSGA-Net work \cite{lu2019nsganet}. However, this approach is tied to the hardware platform the super-network is trained on and would require full training to be redone if the resulting super-network was deployed on a different platform (e.g., trained on GPU, deployed onto Raspberry Pi). Our solution stands outside of the super-network training and fine-tuning mechanics in order to make it generally compatible with any arbitrary super-network framework. It could be jointly used with a NAT approach to inform the training process or used with the popular OFA and HAT approaches to accelerate the post-training sub-network search. 

A successful approach to reduce the search space complexity based on principles introduced by \cite{pmlr-v97-tan19a-efficientnet}, is CompOFA \cite{sahni2021compofa}. CompOFA uses the relationships between elastic parameters to avoid exploring large areas of the model space, significantly improving the time required to find optimal sub-networks. However, the CompOFA approach trades off objective space accuracy due to the search space reduction, requires the supervised application of model architecture-specific expertise and would not translate clearly to other domains.

\section{System Description}

Given the growing popularity of super-network DNN architectures across a plethora of machine learning problem domains, we describe a flexible hardware-aware super-network search system in Figure \ref{fig:system_overview}. For an arbitrary super-network framework and objective space (accuracy, latency, multiply-accumulates (MACs), etc.), our system automates the architecture search process and extracts sub-networks which are optimal for a set of one or more objectives. The system can be interfaced with popular super-network NAS approaches, such as OFA, and can be used in both during training search (e.g., NAT) and post-training search (e.g., HAT) scenarios. The system also works well with off-the-shelf pre-trained super-networks for cases where a user does not have the capability to run training or fine-tuning on their end. Further, if the same super-network topology is used across multiple hardware configurations, the statistical representations of the elastic parameters along with the optimal Pareto front can be used to inform the next sub-network search via warm-start or constrained search approaches which are described in the following sections.

The key sub-network search components of the system are the search tactic library, evaluation manager, and the search module. The search tactic library offers a variety of choices for how to treat the high-level flow of the search such as determining how objectives will be measured (e.g., validation, estimation, proxy) while the other aspect relates to the search flow of the evaluation manager and search module (e.g., full search, concurrent search). For this work, we refer to a measurement of the one or more objectives of a sub-network as an evaluation where a \emph{validation} evaluation is the actual measurement of an objective and a \emph{prediction} evaluation is the predicted measurement of an objective. Additionally, \emph{proxy} scores can be used as a simple heuristic of performance such a model parameter counts or scores such as work by \cite{mellor2021neural}. The evaluation manager handles requests for validation measurement data from the attached super-network framework (e.g., sub-network accuracy and latency) and handles the predictor training and/or look-up table (LUT) construction. The search module handles the system library of single- and multi-objective evolutionary search algorithms (MOEAs) and their associated tuning parameters (e.g., mutation rate, crossover rate, etc.). Other peripheral modules are the analytics module which automates the plotting and descriptive statistics about the search results and the representation module which stores information about past searches and analyses. We describe the details of these system components in more detail in the next section.

\section{Methods}
\label{sec:methods}

In this section we describe the details of the system components and provide the methodologies for the various search tactics. To demonstrate our system on the sub-network search task, we start by showing the use of multi-objective evolutionary algorithm (MOEA) approaches, specifically NSGA-II \cite{deb2002fast}. Next, we highlight the use of warm-start search which works by taking the best sub-network population from a previous NAS run, and using that population as the seed for a different search on a different hardware configuration. Finally, we highlight a novel approach to sub-network search acceleration called \emph{ConcurrentNAS} that leverages the capabilities of weakly trained predictors to minimize the number of validation measurements. Through these examples we demonstrate how our system can accelerate the sub-network search process when interfaced with existing NAS approaches such as OFA and HAT. 

\subsection{Problem Formulation}
\label{ssec:problem_formulation}

\begin{figure}[ht!]
    \centering
    \includegraphics[width=0.5\linewidth]{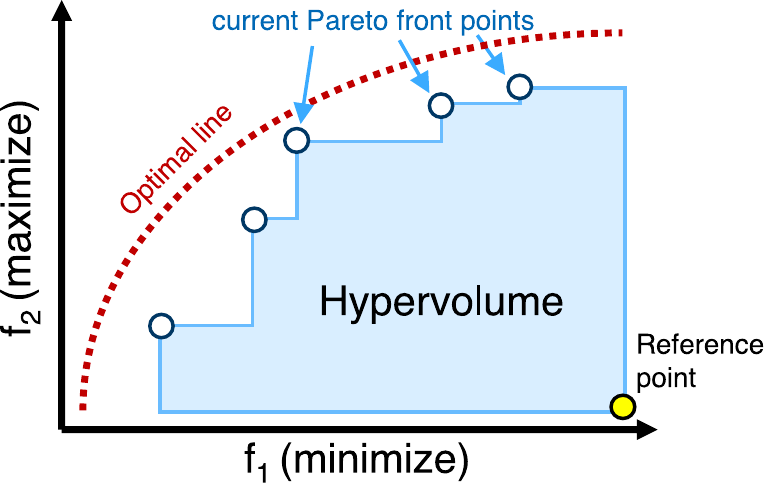}
    \caption{Illustration of a two dimensional objective space with hypervolume and Pareto front points that represent sub-networks in this work.}
    \label{fig:pareto_front_illustration}
\end{figure}

Consider a pre-trained super-network with weights $W$, a set of sub-network architectural configurations $\Omega$ derived from the super-network and $m$ competing objectives $f_1(\omega; W), \ldots, f_m(\omega; W)$ where $\omega \in \Omega$. Each of the sub-network configurations $\omega$ is a valid set of parameters used during training of the super-network. For example, a given $\omega$ will contain values used for each elastic depth and kernel size during super-network training. Our system aims to minimize the objectives $f_i, i \in \{1, \ldots, m\}$ to find an optimal sub-network $\omega^*$. In other words,
\begin{equation}
    \omega^*=\underset{\omega \in \Omega}{\text{argmin}}\big(f_1(\omega; W), \ldots, f_m(\omega; W)\big)
\end{equation}
In the case of maximizing an objective (e.g., accuracy), the objectives can be negated to transform it to a minimization problem. During optimization, multiple $\omega^* \in \Omega$ will be found in different regions of the objective space and form a \textit{Pareto front}. It is this set of optimal sub-networks that are the points on the Pareto front illustrated in Figure \ref{fig:pareto_front_illustration}.

In later comparative studies we use the hypervolume indicator \cite{zitzler1999multiobjective} to measure how well the Pareto front approximates the optimal solution as shown in Figure \ref{fig:pareto_front_illustration}. When measuring two objectives, the hypervolume term represents the dominated \textit{area} of the Pareto front. In the results, we refer to the search solution that has saturated after many evaluations as the \textit{near-optimal} Pareto front.

OFA offers super-networks based on the ResNet50 \cite{he2015deep} and MobileNetV3 \cite{howard2019searching} architectures, while HAT provides a machine translation model that we call the Transformer super-network. The size of the search space of sub-networks derived from the MobileNetV3 super-network is $10^{19}$ and $10^{15}$ for the Transformer super-network.

\subsection{Evolutionary Algorithms}
\label{ssec:evo_algorithms}

In this work we focus on applying multi-objective evolutionary algorithm (MOEA) approaches to the sub-network search problem due to our own success with such approaches and those shown by \cite{Lu_2021_NAT} and \cite{lu2019nsganet}. We choose to use multi-objective (two objectives) over single or many-objective approaches as they are easily interpreted from a Pareto front visual standpoint, but note that our system works with any number of objectives. We limit the MOEA algorithm to NSGA-II for consistency of results and modify the base algorithm to ensure that unused parameters in the search space (e.g., lower block depth can mean other elastic parameters are not used) are accounted for when preventing duplicates. In short, NSGA-II is a \emph{generational} loop process whereby a \emph{population} of \emph{individuals} (sub-networks encoded in terms of the elastic parameters of the super-network) undergo variation via crossover and mutation to create a child population, followed by non-dominated sorting with diversity preservation to select the next generation's population. For this work, we use a mutation rate of $1/population\:size$ and a crossover rate of 0.9. We also note that RNSGA-II \cite{deb2006}, MO-CMA-ES \cite{igel20017}, and AGE-MOEA \cite{panichella2019} have been successfully tested with the system. Furthermore, our system allows for the end-user to apply their own optimization algorithms while still leveraging the other components (e.g., search tactics, evaluation module) of the system. For a detailed survey and overview of evolutionary algorithms, we point the reader to work by \cite{emmerich2018}.

\subsection{Predictors}
\label{ssec:predictors}

Since validation evaluations of performance objectives, such as top-1 accuracy and latency, require a large amount of time, we follow the work in \cite{cai2019once} and \cite{wang2020hat} and employ predictors. More specifically, we predict top-1 accuracy of sub-networks derived from ResNet50 and MobileNetV3 super-networks, BLEU score of sub-networks derived from Transformer super-networks and latency of sub-networks derived from all three super-networks. However, unlike prior work which use multi-layer perceptrons (MLPs) to perform prediction, we employ much simpler methods such as ridge and support vector machine regression (SVR) predictors. The authors of \cite{Lu_2021_NAT} and \cite{anonymous2022what} have found that MLPs are inferior to other methods of prediction for low training example counts. We have found that these simpler methods converge more quickly, require fewer training examples and require much less hyper-parameter optimization than MLPs. The combination of accuracy / latency prediction and using simple predictors allows us to significantly accelerate the selection of sub-networks with minimal prediction error. See Section \ref{ssec:predictor_analysis} for a detailed analysis of our predictor performance.

\subsection{Full Search}
\label{ssec:full_search}

In our system, \emph{full search} means that objective predictors (e.g., accuracy, latency) are first trained from a sampling of measurements (random or supervised) from the super-network architecture space followed by an extensive search using those predictors to estimate architecture performance in the objective space. For example, a full search approach could consist of training accuracy and latency predictors on a GPU platform, and then running a NSGA-II search using these predictors. Another approach is using full search with a warm-start population derived from the most optimal population from another completed search. In our results we show that a full search with warm-start can accelerate the search process by an order of magnitude.  

\subsection{Concurrent Search}
\label{ssec:concurrent_search_methods}

\begin{figure}[tp]
    \centering
     \includegraphics[width=0.6\linewidth]{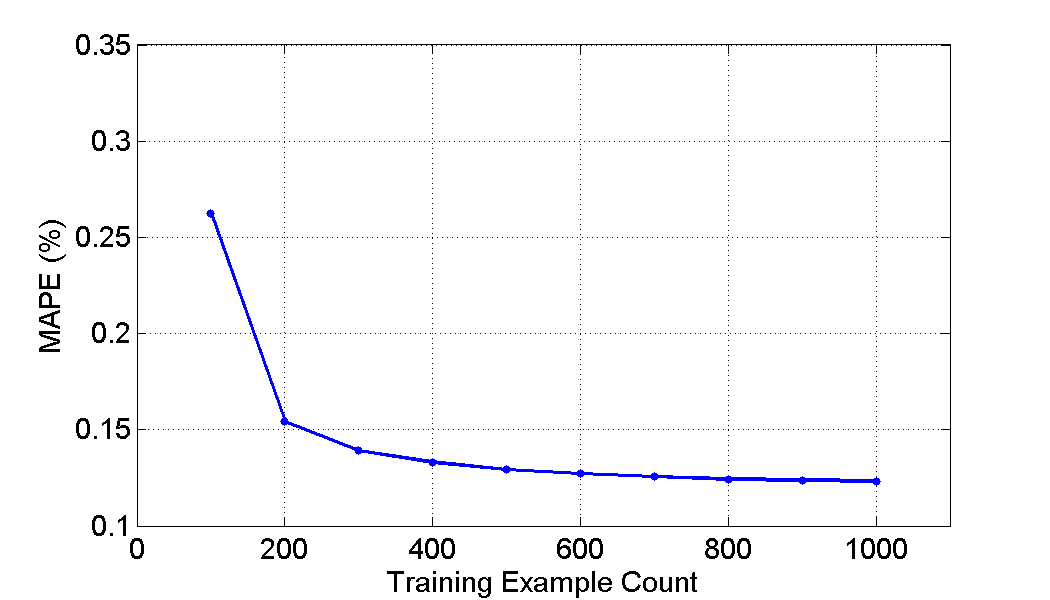}
    \caption{MAPE of our top-1 accuracy predictor versus the number of training examples for sub-networks derived from the MobileNetV3 super-network.}
    \label{fig:mobilenet_accuracy_clx_mape}
\end{figure}

While using trained predictors can speed up the post-training search process, there is still a substantial cost to training predictors as the number of training samples is usually between 1000 and 4000 samples \cite{cai2019once}. In the scenario where compute resources are limited, the goal is to reduce the number of validation measurements as much as possible. As shown in Figure \ref{fig:mobilenet_accuracy_clx_mape}, the accuracy predictor can achieve acceptable mean absolute percentage error (MAPE) in as few as 100 training samples. We build on this insight that weak predictors can offer value during search even when searching holistically across the full Pareto front range. We term this approach concurrent neural architecture search (ConcurrentNAS) since we are iteratively searching with and training a predictor as described in Algorithm \ref{alg:concurrent}. We first take either a randomly sampled or warm-start population to serve as the initial ConcurrentNAS population and then measure each objective score for the individuals (sub-networks) in the population and store the result. The result is combined with all previously saved results and used to train a "weak" predictor. By weak predictor we imply a predictor that has been trained on a relatively small number of training samples. For each iteration, we run a full evolutionary algorithm search (NSGA-II in this work) using the predictor for a high number of generations (e.g., $>200$) to let the algorithm explore the weak predictor objective space sufficiently. Finally, we select the best sub-networks from the evolutionary search to inform the next round of predictor training and so on until the iteration criterion is met or an end-user decides sufficient sub-networks have been identified. We note that a ConcurrentNAS approach can be applied with any single-, multi-, or many-objective evolutionary algorithm and generalizes to work with any super-network framework. Additionally, it provides the flexibility for changing evolutionary parameter tuning parameters as the iterations progress and allows for constraining a subset of objectives to validation evaluations only.

\begin{algorithm}[tb]
   \caption{Concurrent Neural Architecture Search}
   \label{alg:concurrent}
\begin{algorithmic}
   \STATE {\bfseries Input:}
   Objectives $f_m$, super-network with weights $\mathcal{W}$ and configurations $\Omega$, predictor model type for each objective $Y_{m}$, ConcurrentNAS population size $c$, number of ConcurrentNAS iterations $I$, evolutionary algorithm $\mathcal{E}$ with the number of search iterations $J$.
   \STATE \textcolor{gray}{// sample $c$ sub-networks for first population}\\
   \STATE $C_{i=0} \leftarrow \{\omega_{c}\} \in \Omega$ 
   \WHILE{$i < I$}
   \STATE \textcolor{gray}{// measure objectives $f_m$, store results $D_{i,m}$} \\
   \STATE $D_{i,m} \leftarrow f_m(C_{i} \in \Omega; \mathcal{W})$
   \STATE $D_{all,m} \leftarrow D_{all,m} \cup D_{i,m}$
   \STATE $Y_{m,pred} \leftarrow Y_{m,train}(D_{all,m})$ \textcolor{gray}{// train predictors} 
   \WHILE{$j < J$} 
   \STATE $C_{\mathcal{E}_{j}} \leftarrow \mathcal{E}(Y_{m,pred}, j)$ \textcolor{gray}{// run ${\mathcal{E}}$ for $J$ iterations}
   \ENDWHILE
   \STATE $C_{i} \leftarrow C_{\mathcal{E},best} \in C_{\mathcal{E}_{J}}$ \textcolor{gray}{// retrieve optimal population}
   \STATE $i \leftarrow i+1$
   \ENDWHILE
   \STATE {\bfseries Output:} All ConcurrentNAS populations $C_{I}$, search results $C_{\mathcal{E}_{I,J}}$, and validation data $D_{all,m}$. 
   
\end{algorithmic}
\end{algorithm}

\subsection{Population Density-Based Constraints}
\label{ssec:popdb_methods}
In order to minimize the number of samples required to reach the optimal Pareto front during search, we aim to construct a smaller architecture search space $\widetilde{\Omega} \subset \Omega$. We describe here a method for constructing the reduced search space in an \textit{unsupervised} fashion for a family of hardware platforms that share similar characteristics to a single platform which we have already optimized. 

Density-based clustering \cite{ester1996density, campello2013density} groups points that are packed tightly together in the feature space and are assigned a non-negative label. During clustering, these methods mark outlier points as noise.

\begin{figure}[htb]
    \centering 
\includegraphics[width=0.6\linewidth]{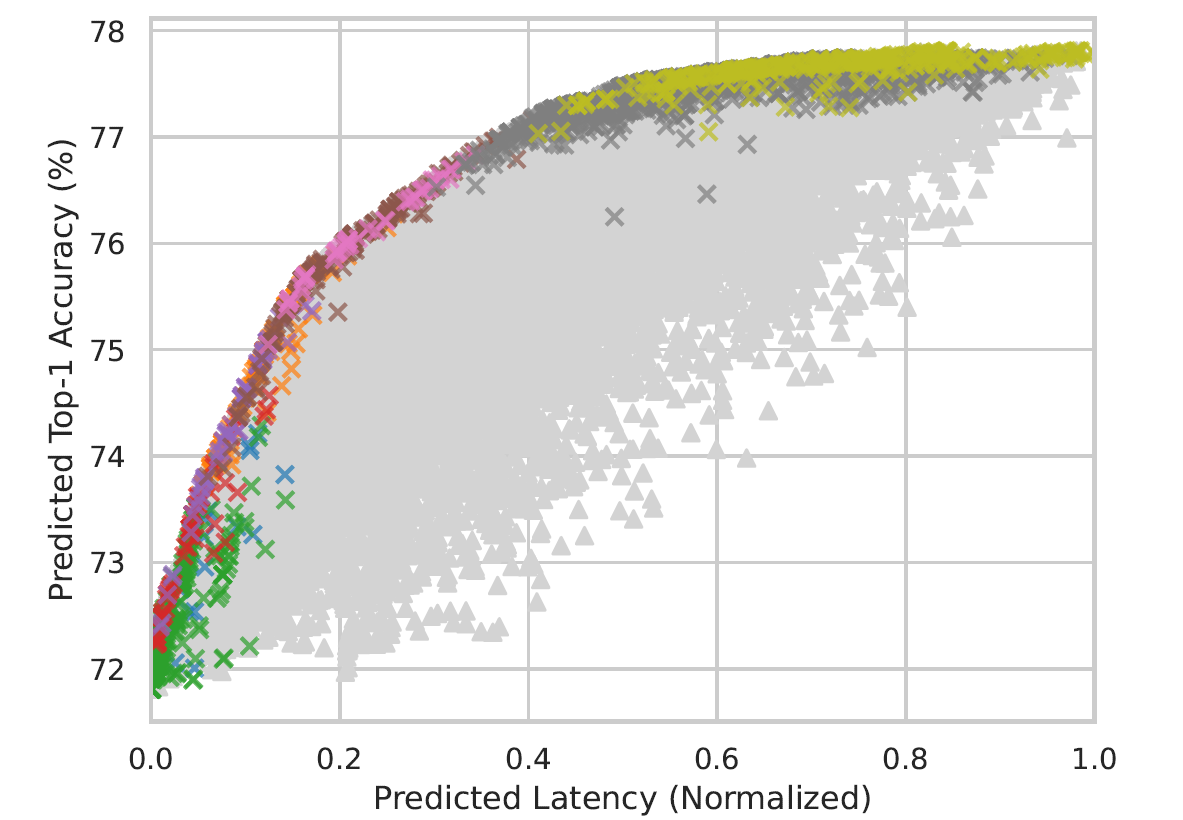}
\caption{Clusters obtained using from running HDBSCAN (\texttt{min\_cluster\_size} = $50$, \texttt{min\_samples} = $10$) on the set of sub-network architectural configurations found through an $100$k sample NSGA-II search. Points marked as colored crosses correspond to configuration that were grouped into clusters by the algorithm, grey triangle points were marked as noise.}
    \label{fig:clustering_anaylsis}
\end{figure}

In practice, clustering using HDBSCAN over the history of full search over a platform will find the heavily explored regions of the search space. This occurs because, due to the evolutionary algorithm, populations will be heavily concentrated around regions of the search space that show good performance along the Pareto front. This tendency to cluster points along the front is shown in Figure~\ref{fig:clustering_anaylsis}.

We compute the relative frequencies of all elastic parameters values belonging to \emph{non-noise} clusters. Then, for every elastic parameter, particular values are excluded from the constrained search space if they do not exceed a threshold defined for that elastic parameter.

We refer to this method of constructing a smaller architecture search space that uses the population densities as \textit{population density-based} (PopDB) constraints.

\subsection{Test Platforms and Considerations}
\label{ssec:platforms_and_considerations}

In this work we use both CPU and GPU platforms for evaluating our system. The hardware platforms and their characteristics are shown in Table \ref{tab:hardware_platforms}.

\begin{table*}[htb]
	\centering
	\begin{tabular}{c|c|c|c}
	\hline \hline
	Name & Memory & 
	\begin{tabular}{@{}c@{}} Thread Count \\ (Host CPU)\end{tabular} &
	\begin{tabular}{@{}c@{}} Microarchitecture \\ (Host CPU)\end{tabular} \\
	\hline
	Intel$\textsuperscript{\textregistered}$ Xeon$\textsuperscript{\textregistered}$ Platinum 8180 & 192 GB & 56 & Skylake (SKX) \\
	Intel$\textsuperscript{\textregistered}$ Xeon$\textsuperscript{\textregistered}$ Platinum 8280 & 192 GB & 56 & Cascade Lake (CLX) \\
	NVIDIA$\textsuperscript{\textregistered}$ Tesla$\textsuperscript{\textregistered}$ V100 & 32 GB & 32 & Skylake (SKX) \\
	NVIDIA$\textsuperscript{\textregistered}$ Tesla$\textsuperscript{\textregistered}$ A100 & 32 GB  & 32 & Cascade Lake (CLX) \\
	\hline \hline
	\end{tabular}
	\caption{Hardware platforms used for system evaluation.}
    \label{tab:hardware_platforms}
\end{table*}

Since our results have latency objective metrics from different manufacturers and there are possible proprietary issues in sharing what could be perceived as official benchmark data, we normalize latencies to be within $[0, 1]$. More specifically, the normalized latency $\hat{l}$ is given by
\begin{equation}
    \hat{l}=\frac{l-l_{min}}{l_{max}} \in [0,1]
\end{equation}
where $l$ is the unnormalized latency, $l_{min}$ is the minimum unnormalized latency and $l_{max}$ is the maximum unnormalized latency. Using normalized latency does not change the underlying search results we are demonstrating. For comparative latency performance metrics related to our test platforms, we point the reader to the MLCommons\footnote{https://mlcommons.org} benchmark suite.

\section{Results}

\begin{figure}[tp]
    \centering
    \includegraphics[width=0.5\linewidth]{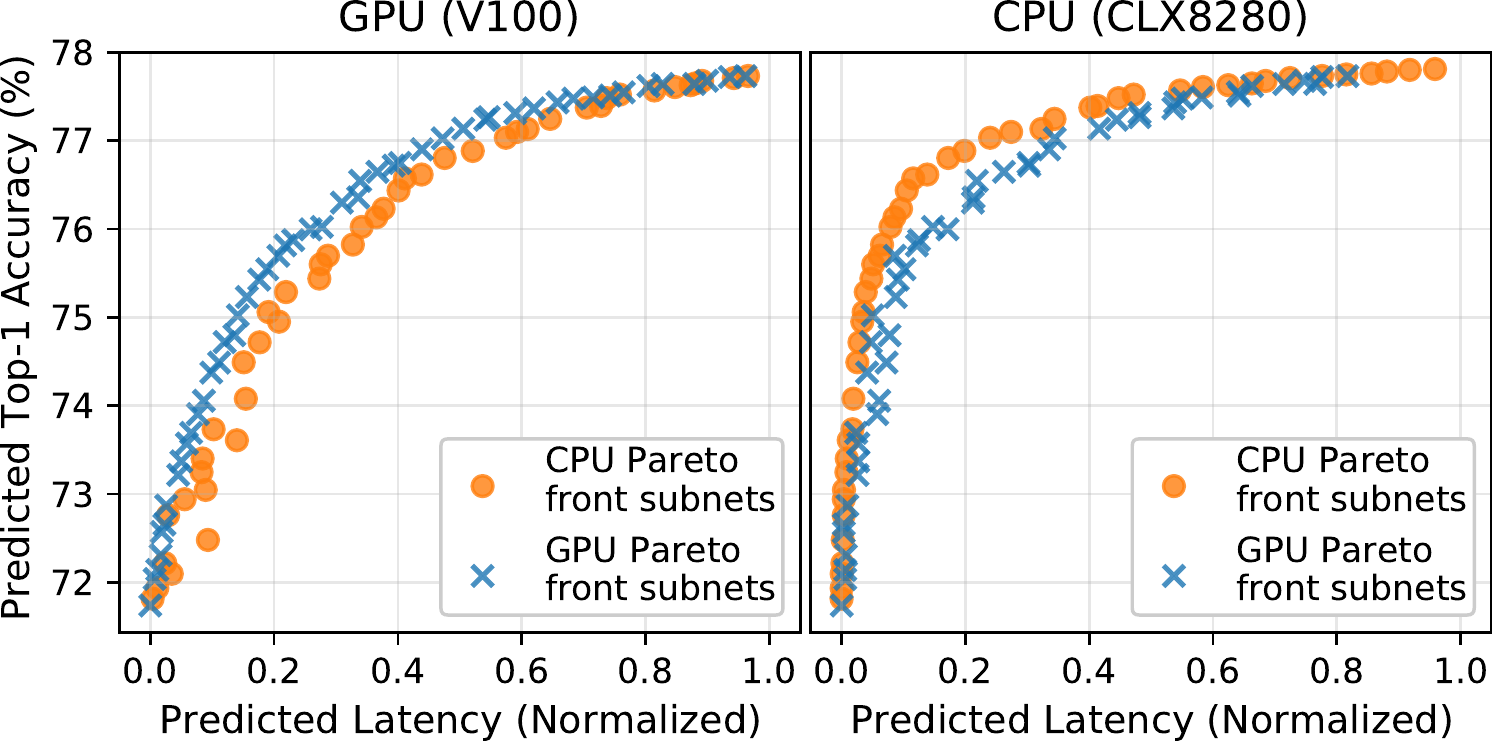}
    \caption{Pareto fronts specialized to GPU (V100) and CPU (CLX) showing that optimal sub-network configurations found on one hardware platform do not translate to the optimal sub-networks for another. Batch size was 128. See Table \ref{tab:hardware_platforms} for details on these hardware platforms.}
    \label{fig:pareto_hw_compare}
\end{figure}

\begin{figure*}[t]
    \centering
    \includegraphics[width=1.0\linewidth]{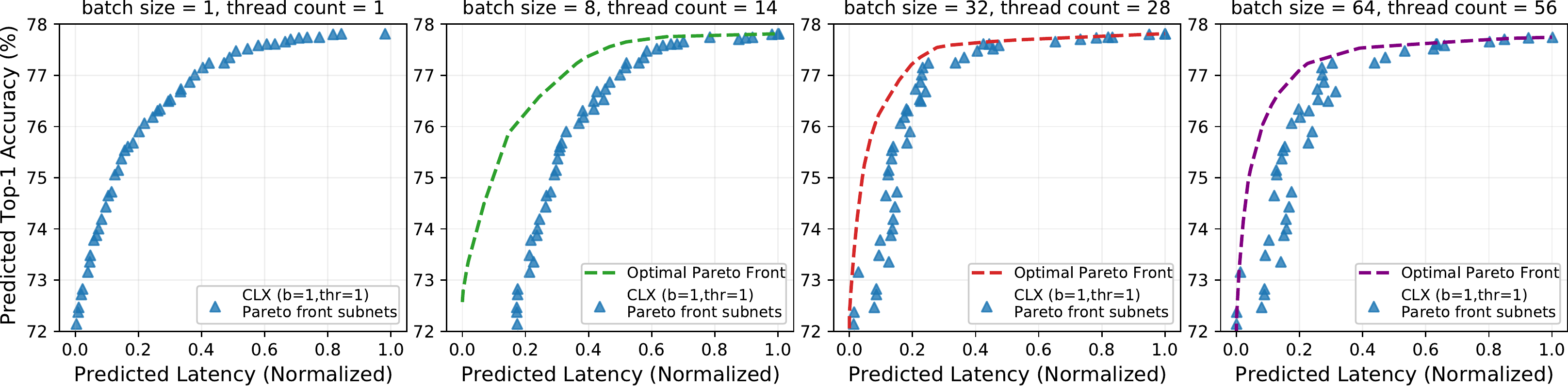}
    \caption{Pareto fronts with CLX for specialized thread counts/batch sizes, and the non-specialized configurations for comparison (MobileNetV3).}
    \label{fig:pareto_clx_configs}
\end{figure*}

A primary goal of our system is to accelerate the sub-network search to address the issue that various hardware platforms and hardware configurations have uniquely optimal sub-networks in their respective super-network parameter search spaces. Figure \ref{fig:pareto_hw_compare} shows that an optimal set of sub-networks found on a CPU platform may not transfer to the optimal objective region on a GPU platform and vice versa. Furthermore, within a hardware platform, Figure \ref{fig:pareto_clx_configs} shows that sub-network configurations found to be optimal to one CPU hardware configuration (e.g., batch size = 1, thread count = 1), do not transfer optimally to other hardware batch size / thread count configurations. 

\subsection{Full Search}
\label{ssec:efficient_search}

\begin{figure*}[htb]
    \centering
    \includegraphics[width=1.0\linewidth]{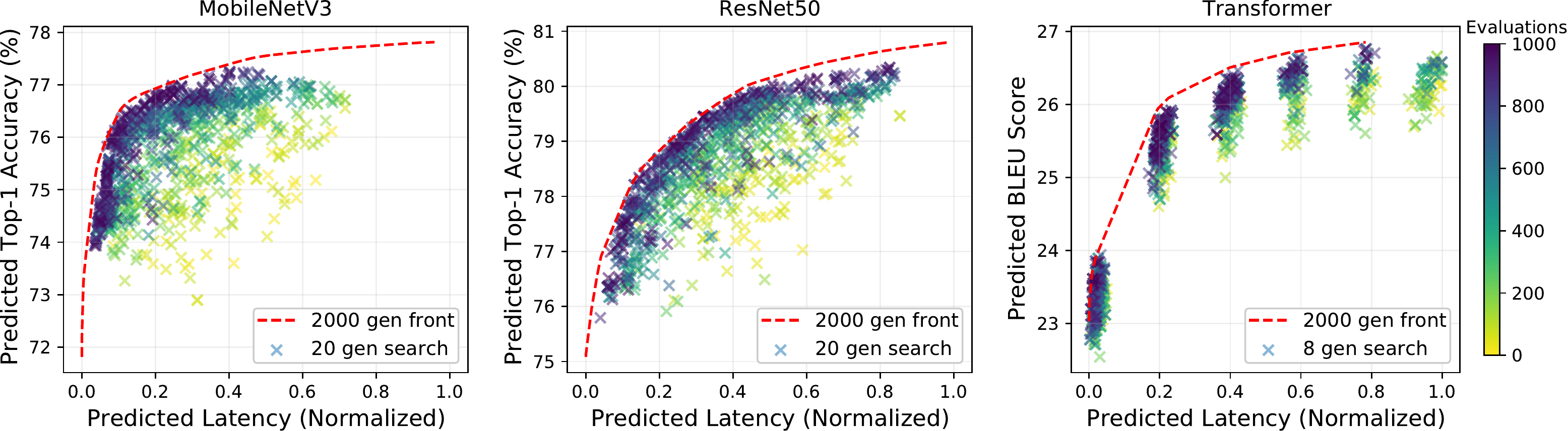}
    \caption{Evolution of the sub-network search for MobileNetV3, ResNet50, and Transformer super-networks on CLX (thread count = 56) using NSGA-II. We use a batch size of 128 for MobileNetV3 and ResNet50 and a batch size of 1 for Transformers. Lighter points are early population generations, darker are from the later generations. The dotted line shows a near-optimal Pareto front from an extended 2000 generation (population size = 50 for ResNet50 and MobileNetV3, population size = 125 for Transformers) search.}
    \label{fig:full_search}
\end{figure*}

The sub-network search progression using our system's modified NSGA-II approach is illustrated across various super-networks as shown in Figure \ref{fig:full_search}. The initial population of the genetic algorithm is randomly sampled from the super-network parameter space as a starting point. As the search progresses through the crossover/mutation and non-dominated diversity preserving sorting, a progression towards the optimal solution region is observed. Both MobileNetV3 and ResNet50 super-networks, being image classification based, show a similar progression in the search whereas the Transformer super-network shows set of discrete clusters that arise due to the different number of decoder layers in the sub-networks. 

\begin{figure}[htb]
    \centering
    \includegraphics[width=0.4\linewidth]{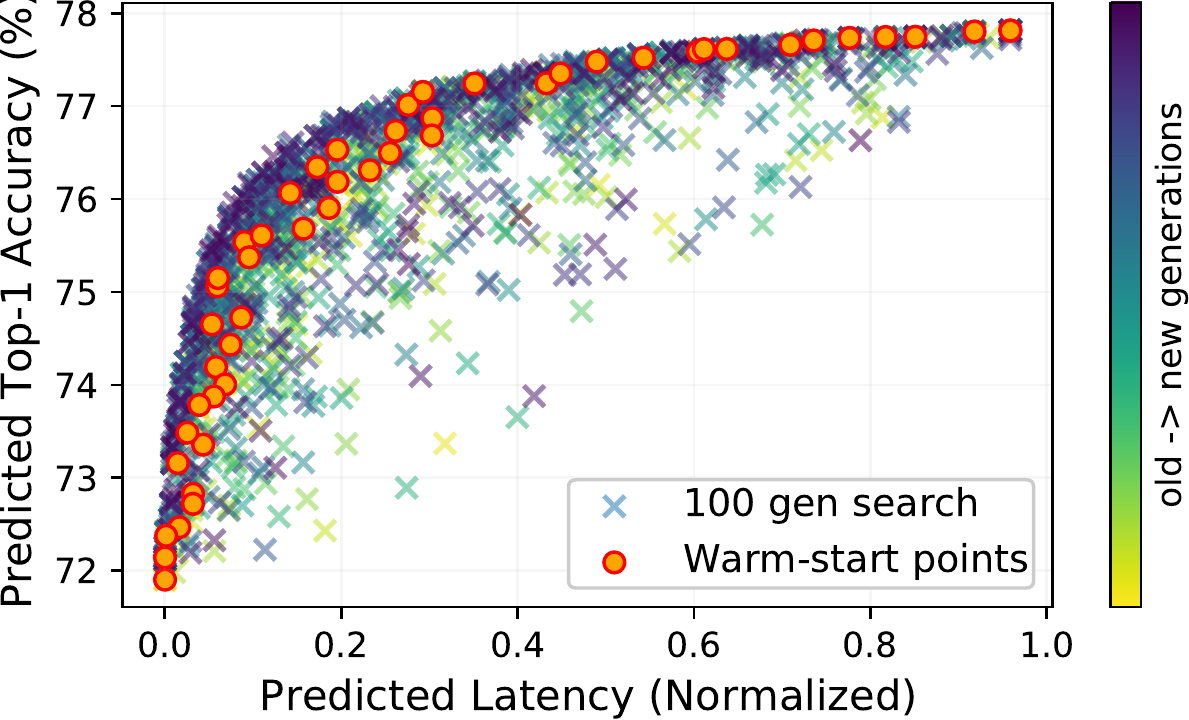}
    \caption{NSGA-II full search showing the evolution of the search for MobileNetV3 using CLX (batch size = 128, thread count = 56) seeded with warm-start points from a CLX (batch size = 1, thread count = 1) result.}
    \label{fig:warm_start}
\end{figure}

One insight from Figure \ref{fig:pareto_clx_configs} is that while the "best" sub-networks found on one hardware configuration do not always translate to the near-optimal Pareto region for another, they could be used as an initial population for a new search on a different hardware setup. The transferable population or "warm-start" approach is shown in Figure \ref{fig:warm_start}. The biggest advantage of warm-start is that it gives a starting population that spans the full range of the objective space region and thus introduces a diverse and better performing population at the start. Due to the crossover/mutation mechanics of the genetic algorithm search, there are still sub-networks that are sampled behind the warm-start front, but the density of those samples is far less than a randomly initialized NSGA-II search.

\subsection{Concurrent Search}
\label{ssec:concurrent_search_res}

\begin{figure}[htb]
    \centering
    \includegraphics[width=0.8\linewidth]{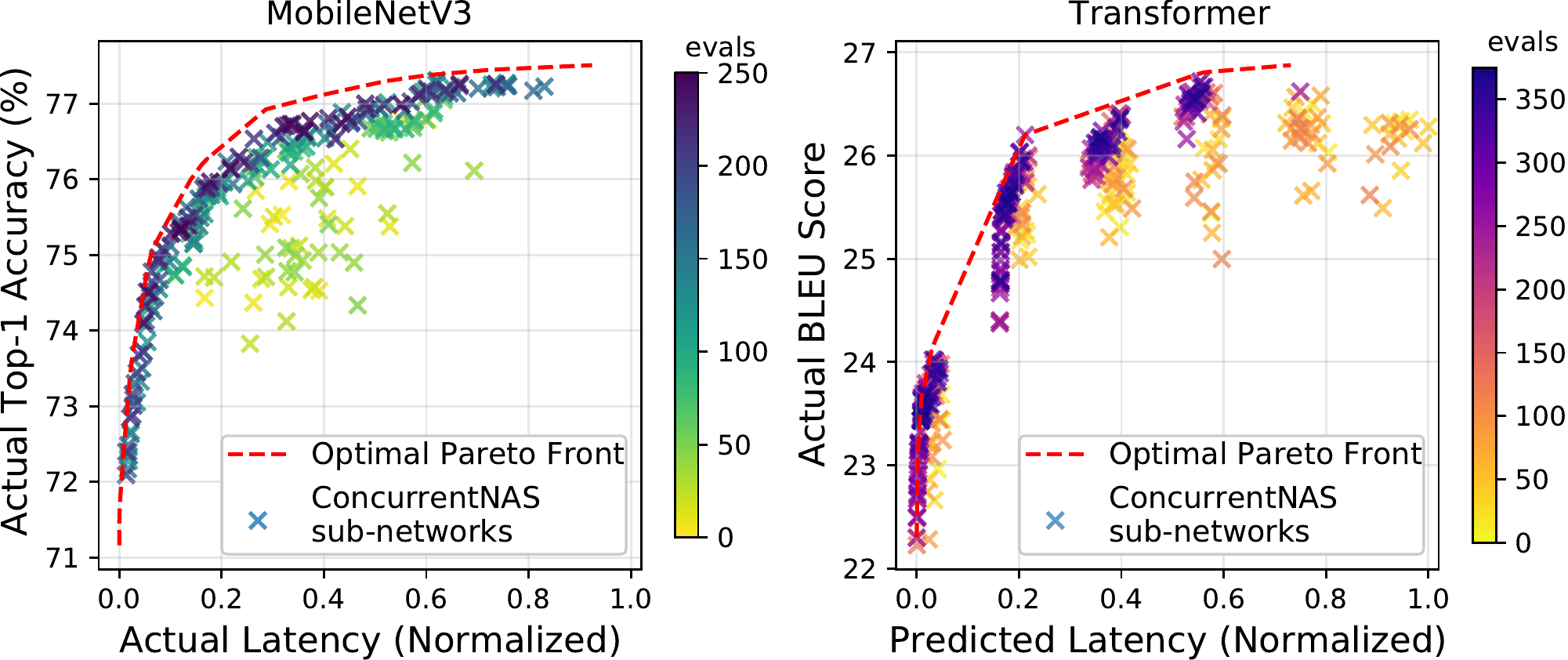}
    \caption{ConcurrentNAS performance on MobileNetV3 (5 generations with population size = 50) and Transformer (4 generations with population size = 125) super-networks on CLX highlighting the fast acceleration towards the optimal Pareto front with minimal validation evaluations.}
    \label{fig:concurrent_mbnv3}
\end{figure}

\begin{figure}[htb]
    \centering
    \includegraphics[width=0.6\linewidth]{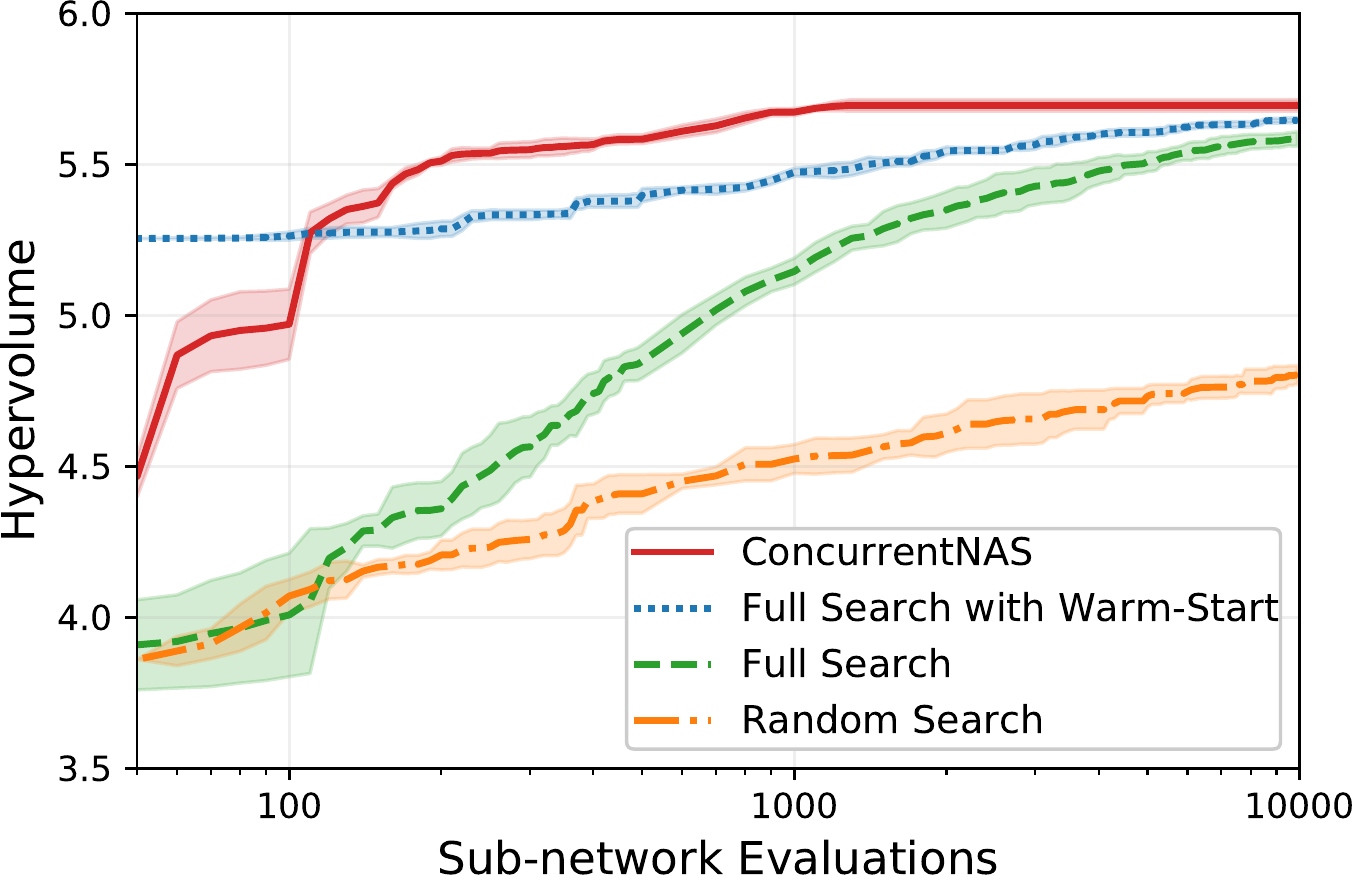}
    \caption{Hypervolume comparison on CLX (batch size 128) for different search tactics in our system that shows how rapidly ConcurrentNAS accelerates to the optimal Pareto Front. The shaded regions represent the standard deviation for 5 trials.}
    \label{fig:hypervolume}
\end{figure}

The main goal of ConcurrentNAS is to reduce the total number of validation measurements required to find optimal model architectures for a given super-network. Additionally, a ConcurrentNAS approach offers the benefit of providing the actual validation population results. Figure \ref{fig:concurrent_mbnv3} shows the evolution of the concurrent search of sub-networks derived from MobileNetV3 and Transformer super-networks respectively where as opposed to Figure \ref{fig:full_search}, the sub-network search accelerates towards the near-optimal Pareto front in only a few generations. The ConcurrentNAS validation measurement cost in evaluations versus hypervolume when compared across approaches is shown in Figure \ref{fig:hypervolume}. As expected a full validation-only search approach with a warm-start population immediately translates to a larger hypervolume in early evaluations. A full search with 10,000 evaluations or more would eventually catch up to the warm-start approach. A more important observation is how quickly ConcurrentNAS overtakes full search, since each concurrent validation population represents the best "learned" objective space information from the predictors. This highlights how rapidly sub-network search can be accelerated and the usefulness of ConcurrentNAS when optimizing for many hardware platforms and configurations. 

\begin{figure}[htb]
    \centering
    \includegraphics[width=0.6\linewidth]{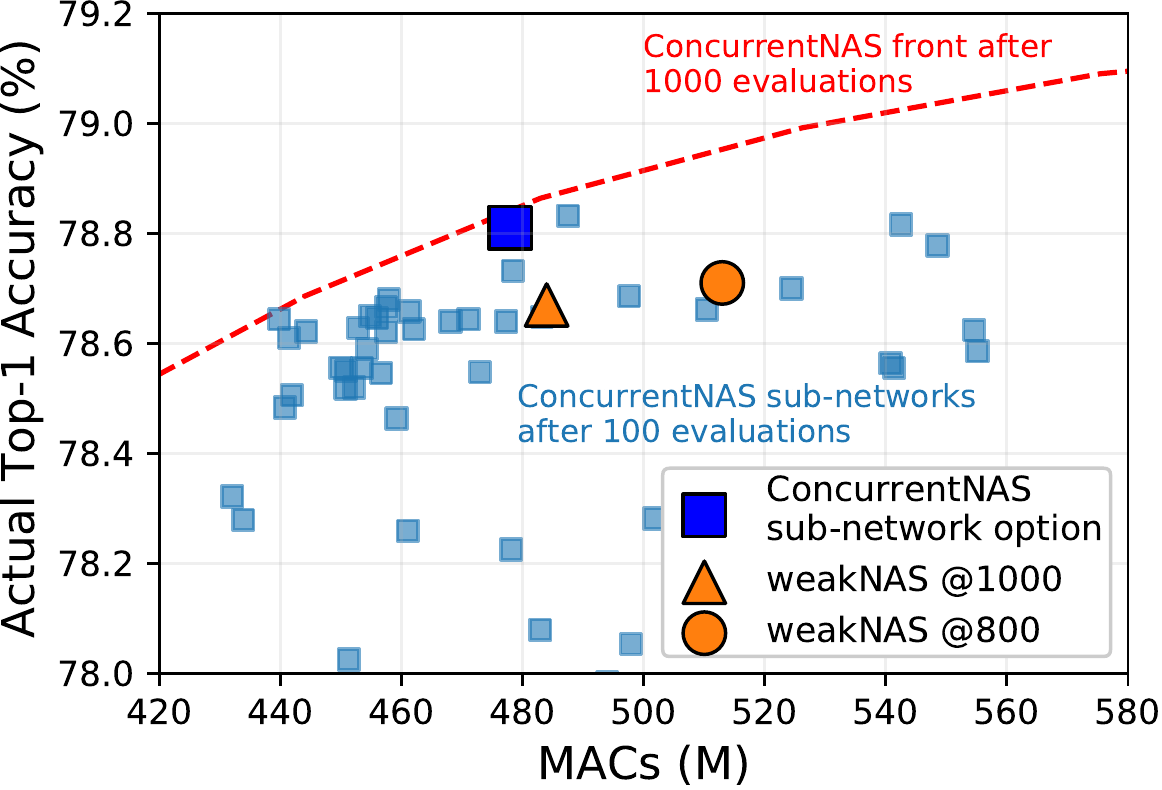}
    \caption{Comparison of ConcurrentNAS versus WeakNAS on the MobileNetV3-w1.2 pre-trained super-network. ConcurrentNAS is able to identify competitive sub-networks (blue squares) in as few a 100 validation measurements.}
    \label{fig:bench_weaknas}
    
\end{figure}

For a comparative benchmark, we evaluate the total number of validation evaluations needed by ConcurrentNAS against those of WeakNAS \cite{wu2021stronger} for the pre-trained OFA MobileNetV3-w1.2 super-network. Figure \ref{fig:bench_weaknas} shows the sub-networks identified in 800 and 1000 evaluations by WeakNAS versus those found by ConcurrentNAS in only 100 evaluations. Table \ref{tab:weaknas_compare} shows the post-search (but prior to fine-tuning) metrics for validation evaluations, top-1 accuracy, and MACs between approaches. We limit the comparison in Table \ref{tab:weaknas_compare} to the pre-trained OFA MobileNetV3-w1.2 super-network that was used by WeakNAS. It is important to note that state-of-the-art accuracy results shown by WeakNAS and OFA represent specific \emph{post-search} fine-tuning approaches that increase the accuracy of the identified sub-networks. 

\begin{table}[htb]
    \centering
    \begin{tabular}{cccc}
    \hline \hline
    \multicolumn{1}{c}{Approach} &
    \multicolumn{1}{c}{\begin{tabular}[c]{@{}c@{}}Validation\\Count\end{tabular}} & \multicolumn{1}{c}{\begin{tabular}[c]{@{}c@{}}Top-1\\(\%)\end{tabular}} & \multicolumn{1}{c}{\begin{tabular}[c]{@{}c@{}}MACs\\(M)\end{tabular}} \\
    \hline
    WeakNAS & 1000 & 78.71 & 484 \\
    WeakNAS & 800 & 78.70 & 513 \\
    ConcurrentNAS & \textbf{100} & \textbf{78.81}  & \textbf{478} \\
    \hline \hline
    \end{tabular}
    \caption{Comparison of NAS models for the OFA pre-trained MobileNetV3-w1.2 super-network search associated with Figure \ref{fig:bench_weaknas}. ConcurrentNAS finds competitive sub-networks in only 100 validations.}
    \label{tab:weaknas_compare}
\end{table}

\subsection{PopDB Search Space Reduction}
\label{ssec:popdb_res}

\begin{figure}[htb]
    \centering
     \includegraphics[width=0.6\linewidth]{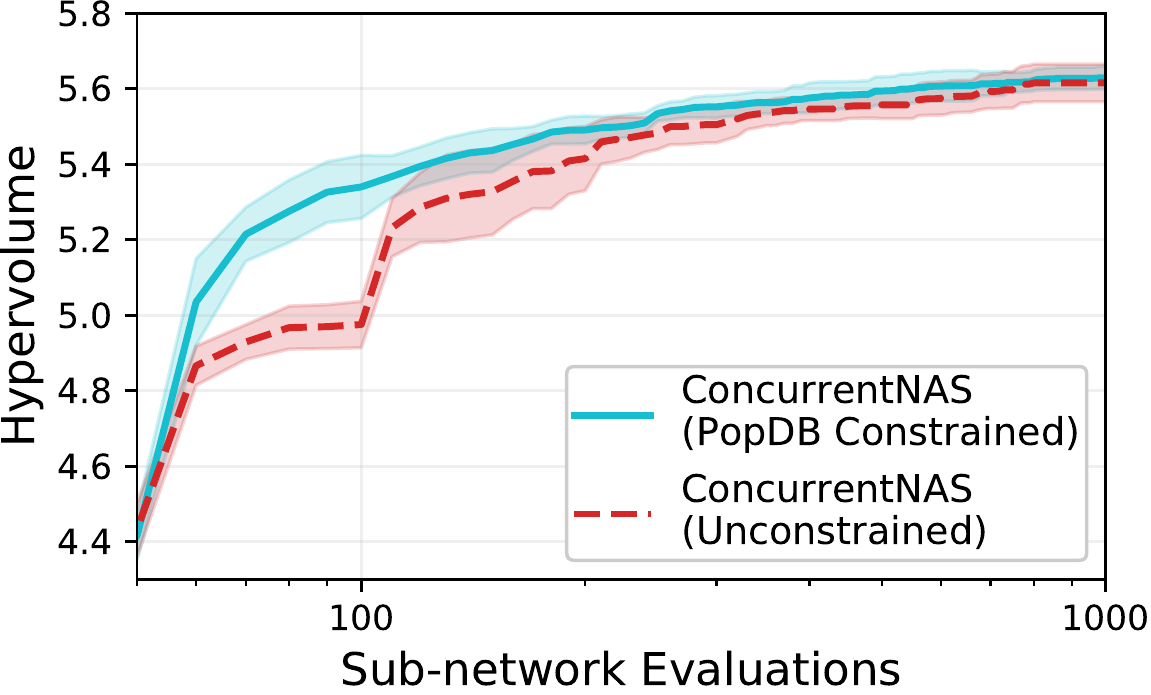}
    \caption{Concurrent search comparison for PopDB constrained search space with size $10^{15}$ versus unconstrained search space $10^{19}$ using CLX (batch size = 256). The shaded regions represent the standard deviation for 5 trials.}
    \label{fig:concurrent_popdb}
\end{figure}

In Figure \ref{fig:concurrent_popdb} we show the measured hypervolume of a ConcurrentNAS search on MobileNetV3 using a constrained search space $\widetilde{\Omega}$ with size $\approx10^{15}$ (see Section \ref{ssec:popdb_methods}). The threshold was set to $1\%$ for all elastic parameters. With this threshold $16$ out of $20$ kernel size, $5$ out of $20$ width, and $0$ out of $5$ depth elastic parameters had values eliminated from the search space. This suggests that kernel sizes of the MobileNetV3 super-network is more invariant across platforms than either depth or width. We observed the effect of a constrained search space as a faster increase in hypervolume during search. Both constrained and unconstrained searches converge to close to the same maximum hypervolume, and that this is consistent across the hardware platforms we considered.

\subsection{Predictor Analysis}
\label{ssec:predictor_analysis}

As described in Section \ref{ssec:predictors}, our work makes extensive use of predictors to accelerate the selection of sub-networks, particularly in the case of concurrent search (see Section \ref{ssec:concurrent_search_methods}). Predictors are necessary since performing actual measurements of metrics such as accuracy or latency would be prohibitively slow. In light of their importance, a better understanding of their performance is needed.

The analysis of the predictors is performed over a number of different trials to account for variance in the results. In each trial, the data sets for each predictor are first split into train and test sets. Subsets of the train data set with 100 to 1000 examples are used to train the predictor. For a given trial, the \textit{same} test set with 500 examples is then used to compute the prediction mean absolute percentage error (MAPE). This process is repeated for a total of 10 trials.

\subsubsection{Accuracy Prediction}
\label{ssec:accuracy_prediction}

\begin{figure*}[htb]
    \centering
    \begin{subfigure}{0.33\textwidth}
      \centering
      \includegraphics[width=1.0\linewidth]{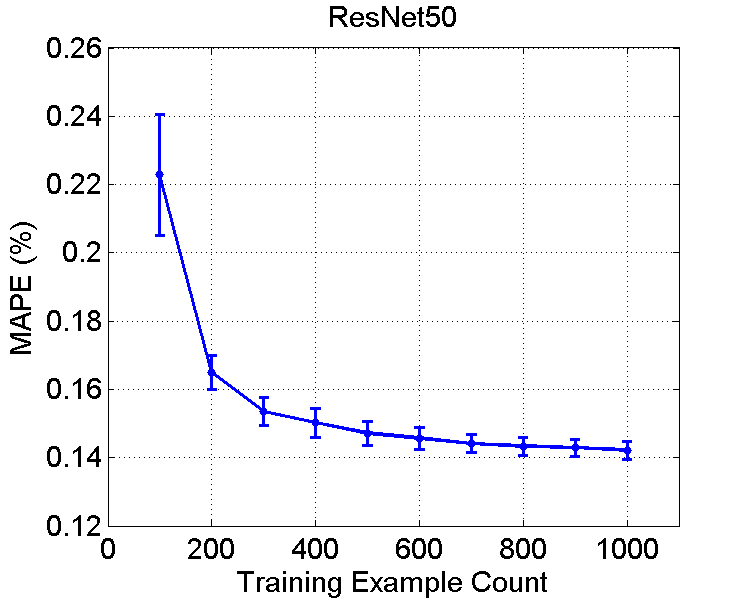}
    \end{subfigure}%
    \begin{subfigure}{0.33\textwidth}
      \centering
      \includegraphics[width=1.0\linewidth]{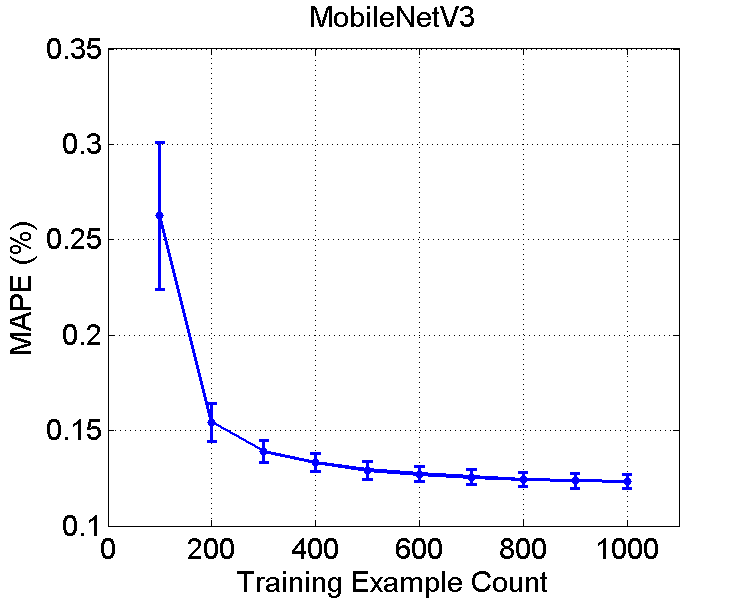}
    \end{subfigure}
    \begin{subfigure}{0.33\textwidth}
      \centering
      \includegraphics[width=1.0\linewidth]{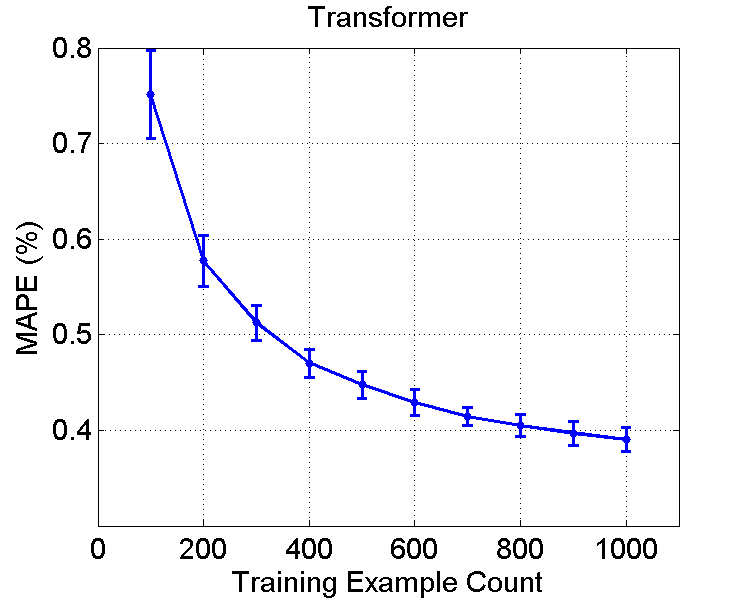}
    \end{subfigure}
    \caption{\label{fig:predictor_analysis_accuracy_mape}MAPE of the ridge predictor performing top-1 accuracy prediction versus the number of training examples for sub-networks derived from ResNet50 (left) and MobileNetV3 (center). MAPE of the SVR predictor performing BLEU score prediction versus the number of training examples for sub-networks derived from Transformer (right). Each point is the average over 10 trials with error bars showing one standard deviation.}
\end{figure*}

We use a ridge predictor to perform top-1 accuracy prediction for sub-networks derived from ResNet50 and MobileNetV3 super-networks and a support vector machine regression (SVR) predictor to perform bilingual evaluation understudy (BLEU) \cite{papinenibleu2002} score prediction for sub-networks derived from the Transformer super-network. The MAPE of these predictors as a function of the number of training examples is shown in Figure \ref{fig:predictor_analysis_accuracy_mape}. In each case, the predictors demonstrate low MAPE for small training example counts and fast convergence.

\begin{figure*}[htb]
    \centering
    \begin{subfigure}{0.33\textwidth}
      \centering
      \includegraphics[width=1.0\linewidth]{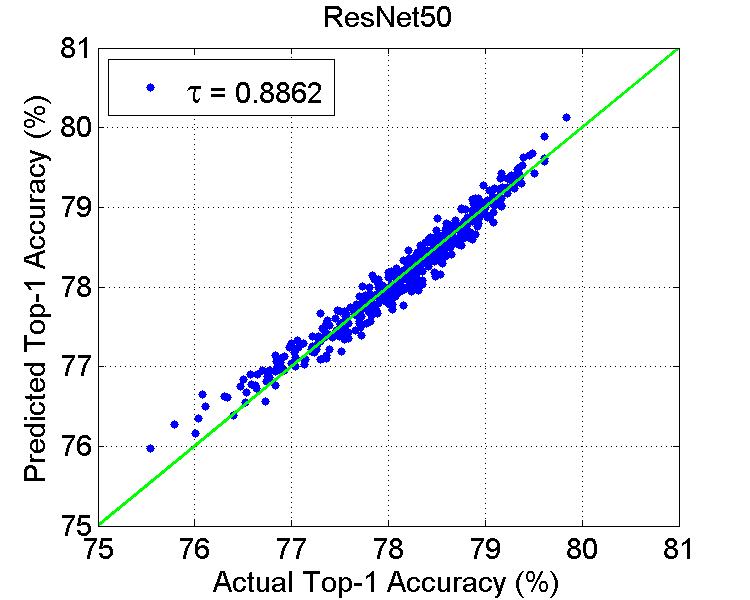}
    \end{subfigure}%
    \begin{subfigure}{0.33\textwidth}
      \centering
      \includegraphics[width=1.0\linewidth]{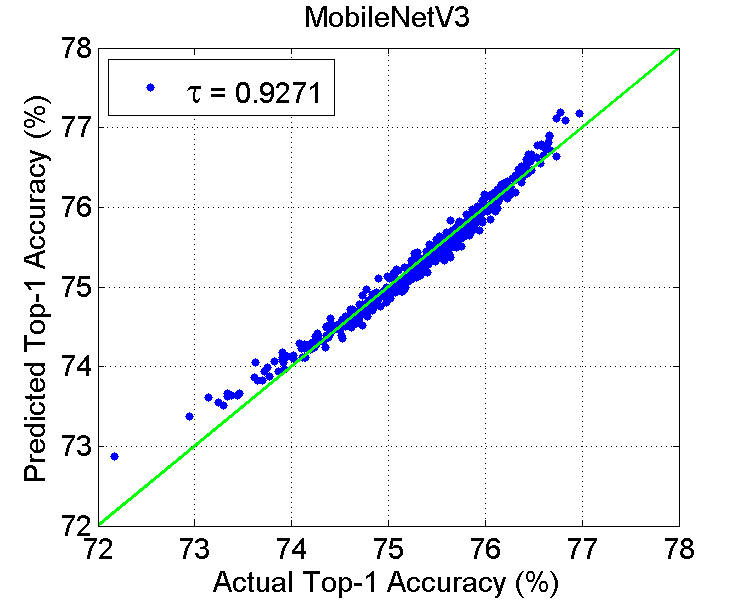}
    \end{subfigure}%
    \begin{subfigure}{0.33\textwidth}
      \centering
      \includegraphics[width=1.0\linewidth]{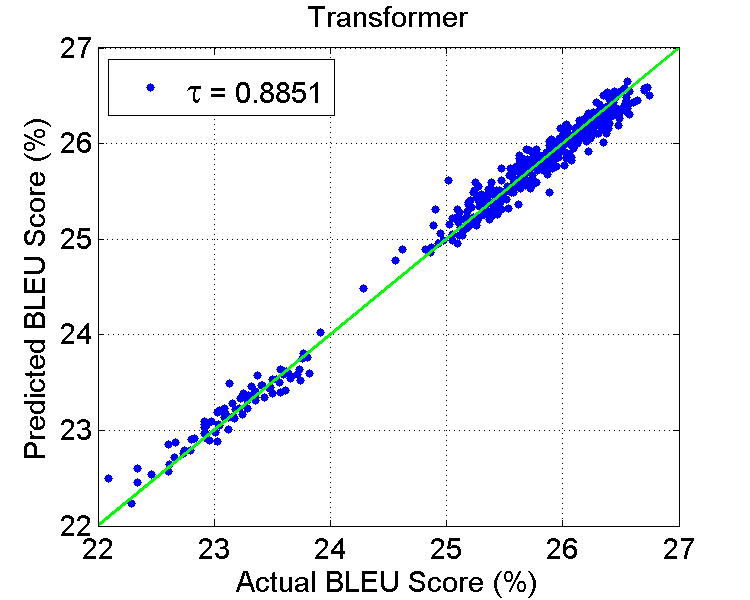}
    \end{subfigure}
    \caption{\label{fig:predictor_analysis_accuracy_correlation}Correlation between ResNet50-derived (left) and MobileNetV3-derived (center) top-1 accuracies and the corresponding predictions and between Transformer-derived (right) BLEU scores and the corresponding predictions. The ideal correlation is shown by the green line.}
\end{figure*}

To visualize accuracy of the predictors, the correlation between actual and predicted top-1 accuracies / BLEU scores are shown in Figure \ref{fig:predictor_analysis_accuracy_correlation}. Along with each correlation is the associated Kendall rank correlation coefficient $\tau$. The actual and predicted accuracies are highly correlated and distributed tightly around the ideal correlation line.

\subsubsection{Latency Prediction}
\label{ssec:latency_prediction}

Not only do we predict accuracy to accelerate the sub-network search, we employ \textit{latency} prediction to further decrease search time. Latency prediction is preferable to LUTs, since a given LUT will be highly specialized to a hardware platform / configuration. Also, trained predictors are, in general, more tolerant to noise in the measurements.

\begin{figure*}[htb]
    \centering
    \begin{subfigure}{0.33\textwidth}
      \centering
      \includegraphics[width=1.0\linewidth]{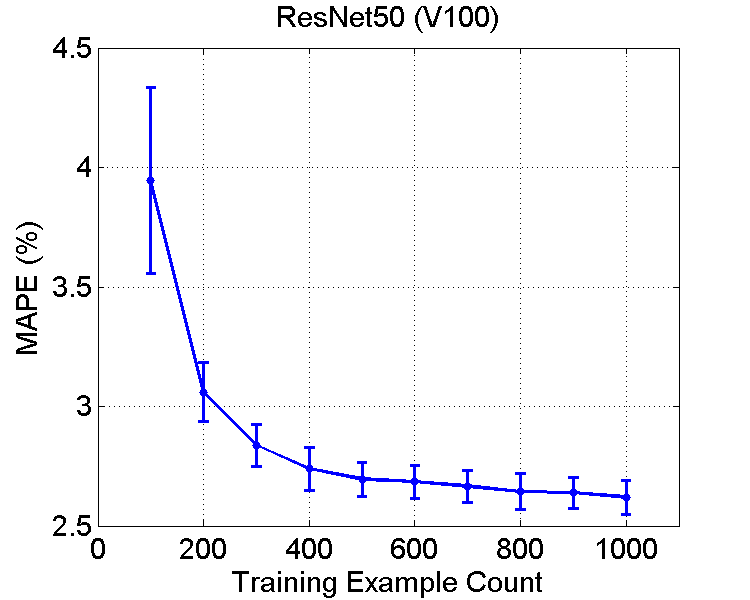}
    \end{subfigure}%
    \begin{subfigure}{0.33\textwidth}
      \centering
      \includegraphics[width=1.0\linewidth]{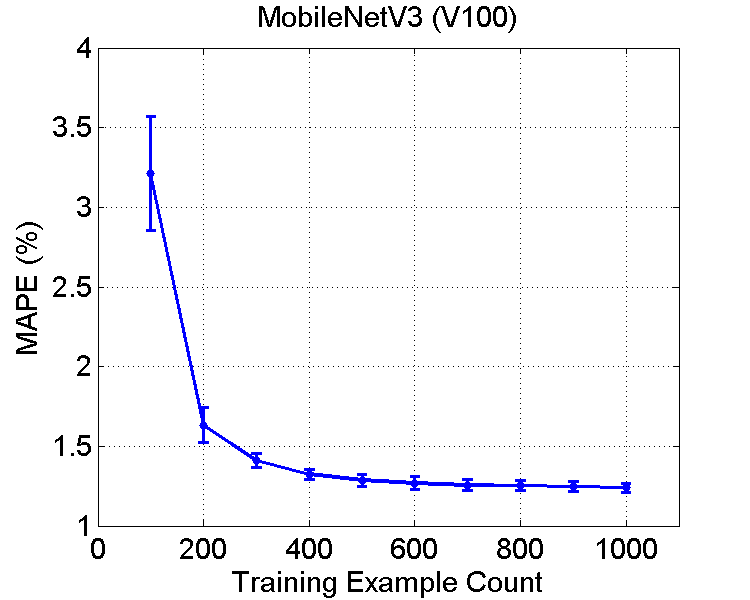}
    \end{subfigure}%
    \begin{subfigure}{0.33\textwidth}
      \centering
      \includegraphics[width=1.0\linewidth]{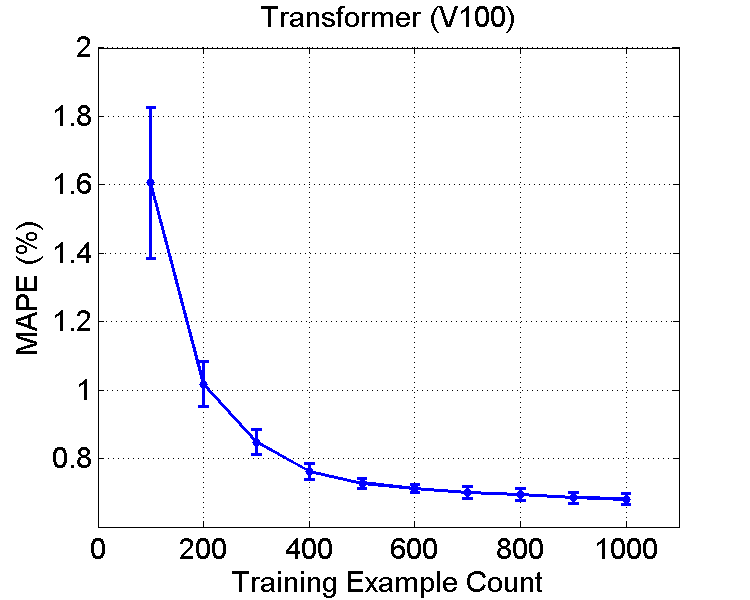}
    \end{subfigure}
    \caption{\label{fig:predictor_analysis_latency_mape}MAPE of the ridge predictors used to perform latency prediction on an NVIDIA V100 GPU versus the number of training examples for sub-networks derived from ResNet50 (left), MobileNetV3 (center) and Transformer (right). Each point is the average over 10 trials with error bars showing one standard deviation.}
\end{figure*}

Similar to the analysis described in Section \ref{ssec:accuracy_prediction}, we use a ridge predictor to perform latency prediction for sub-networks derived from ResNet50, MobileNetV3 and Transformer super-networks. The MAPE of the predictor for an NVIDIA V100 as a function of the number training examples is shown in Figure \ref{fig:predictor_analysis_latency_mape}. Much like the results for accuracy prediction, the latency predictor demonstrates both fast convergence and low MAPE for few training samples.

The $\tau$ values between actual and predicted latencies of sub-networks derived from ResNet50, MobileNetV3 and Transformer super-networks along with different hardware platforms are shown in Table \ref{tab:predictor_analysis_latency_correlation_coefficients}. The latency predictor is highly accurate and produces $\tau$ values greater than 0.86 for all combinations.

\begin{table}[htb]
	\centering
	\begin{tabular}{c|c|c|c|c}
	\hline \hline
	 & SKX & CLX & V100 & A100 \\
	\hline
	ResNet50 & 0.8984 & 0.8690 & 0.8675 & 0.8637 \\
	MobileNet & 0.9821 & 0.9826 & 0.9311 & 0.9360 \\
	Transformer & 0.9206 & 0.8999 & 0.9180 & 0.8838 \\
	\hline \hline
	\end{tabular}
	\caption{Kendall rank correlation coefficient $\tau$ for latency prediction with different networks on different hardware platforms. We compute $\tau$ over 500 different sub-networks for each super-network / hardware platform combination. See Table \ref{tab:hardware_platforms} for details on these hardware platforms.}
    \label{tab:predictor_analysis_latency_correlation_coefficients}
\end{table}

\section{Conclusion}
We have proposed and demonstrated a comprehensive system that more efficiently finds a diverse set of sub-networks from a pre-trained super-network than prior methods. By applying our system to a multi-objective (e.g., latency \textit{and} accuracy) performance space, we have significantly accelerated the search of these sub-networks using novel tactics and algorithms and highly accurate objective predictors. Further, we have shown that even for a given hardware platform (e.g., CPU or GPU), the hardware configuration (e.g., number of threads) must be considered during the search / optimization process. Our approach does not require the super-network to be refined for the target task a priori. We have shown that it interfaces seamlessly with state-of-the-art super-network training methods OFA and HAT. Finally, we have shown that sub-network search is 8x faster with ConcurrentNAS than state-of-the-art WeakNAS approach, and we further accelerate the search with our proposed unsupervised search space reduction technique PopDB.

\bibliographystyle{unsrt}  
\bibliography{paper}

\end{document}